\def\year{2020}\relax
\documentclass[letterpaper]{article} 
\usepackage{aaai20}  
\usepackage{times}  
\usepackage{helvet} 
\usepackage{courier}  
\usepackage[hyphens]{url}  
\usepackage{graphicx} 
\usepackage{graphicx}  
\usepackage{amsfonts}
\usepackage{graphicx}
\usepackage{subfigure}
\usepackage{booktabs}
\usepackage{amssymb}
\usepackage{pifont}
\usepackage{amsmath}
\usepackage{dsfont}
\usepackage{mathrsfs}
\usepackage{mathtools}
\usepackage{makecell}
\usepackage{graphbox}

\usepackage{caption}
\usepackage{subfig}

\title{Learning General Latent-Variable Graphical Models with\\Predictive Belief Propagation}

\author{Borui Wang\\Computer Science Department\\Stanford University\\wbr@cs.stanford.edu\\
\And Geoffrey Gordon\\Machine Learning Department\\Carnegie Mellon University\\ggordon@cs.cmu.edu}

\makeatletter
\newcommand*{\centerfloat}{%
  \parindent \z@
  \leftskip \z@ \@plus 1fil \@minus \textwidth
  \rightskip\leftskip
  \parfillskip \z@skip}
\makeatother

\newcommand{\CI}{\mathrel{\perp\mspace{-10mu}\perp}}

\newcommand\addtag{\refstepcounter{equation}\tag{\theequation}}

\begin{document}

\maketitle

\begin{abstract}
Learning general latent-variable probabilistic graphical models is a key theoretical challenge in machine learning and artificial intelligence. All previous methods, including the EM algorithm and the spectral algorithms, face severe limitations that largely restrict their applicability and affect their performance. In order to overcome these limitations, in this paper we introduce a novel formulation of message-passing inference over junction trees named predictive belief propagation, and propose a new learning and inference algorithm for general latent-variable graphical models based on this formulation. Our proposed algorithm reduces the hard parameter learning problem into a sequence of supervised learning problems, and unifies the learning of different kinds of latent graphical models into a single learning framework, which is local-optima-free and statistically consistent. We then give a proof of the correctness of our algorithm and show in experiments on both synthetic and real datasets that our algorithm significantly outperforms both the EM algorithm and the spectral algorithm while also being orders of magnitude faster to compute.
\end{abstract}

\section{Introduction}
Probabilistic graphical models with latent variables are powerful in modeling many important problems in machine learning and artificial intelligence \cite{Koller09,Bishop06}. The existence of latent variables in such models provides us with the ability to capture richer statistical dependencies among observed variables. However, learning latent-variable graphical models is often difficult, due to the non-convex nature of their parameter learning problem and the intractability in the corresponding inference procedure.

Currently, there are two main types of learning algorithms for latent-variable graphical models. The first one is the Expectation-Maximization algorithm \cite{Dempster77}, which transforms the parameter learning problem into an iterative procedure that maximizes a non-convex objective function. However, the EM algorithm only provides weak theoretical guarantees, can lead to bad local optima and has slow convergence. To address these problems of EM, recently a second type of methods called spectral learning has been proposed \cite{Hsu09,Anima14,Parikh11}. Spectral learning algorithms are based on the idea of method of moments and reparametrize the latent graphical models such that the learning procedure can be performed through tensor algebra using only observed quantities. Although spectral algorithms enjoy the benefits of being provably consistent, computationally efficient and local-optima-free, they have three key limitations. First, most spectral learning algorithms only apply to restricted types of latent structures (mostly trees), and are hard to generalize to more complicated latent structures beyond trees (e.g. loopy graphs). Second, most spectral learning algorithms can only deal with discrete random variables and cannot be easily extended to handle continuous random variables. Third, the current spectral algorithms are generally idiosyncratic to the specific model structures that they are targeted to learn, and thus cannot provide a flexible learning framework to incorporate different prior knowledge and probabilistic assumptions when facing different learning scenarios.

\begin{figure}
\centerline{\includegraphics[width=0.95\columnwidth]{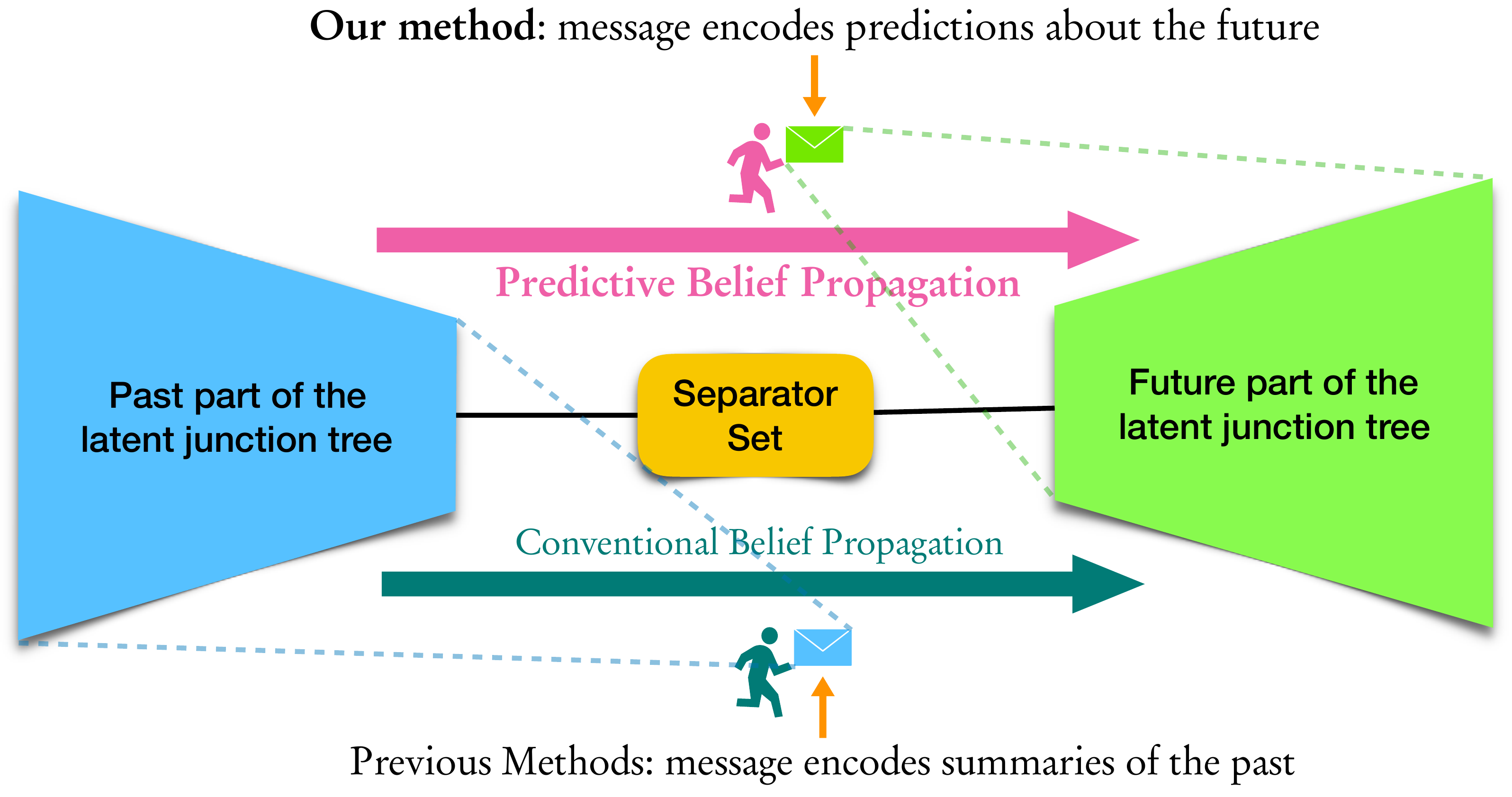}}
\caption{Comparison between predictive belief propagation and conventional methods for belief propagation during inference over latent-variable junction trees.}
\end{figure}

In order to overcome these limitations of previous methods, in this paper we propose a new algorithm for learning general latent-variable graphical models that applies to all different types of latent structures, can handle both discrete and continuous variables of arbitrary forms of probability distribution in a nonparametric fashion, and allows us to incorporate different types of prior knowledge into the learning process, while still remaining provably consistent, local-optima-free and fast to compute. To achieve this, we introduce a new way of formulating message-passing inference over junction trees called \textbf{predictive belief propagation}. In contrast to conventional formulations of message passing, which treat a message as a direct summary of all the probabilistic information seen in the past, we instead think of a message as encoding our predictions about the probabilistic information of all the variables in the future part of the graphical model given what we have seen in the past. This new perspective allows us to systematically reparametrize message passing inference on latent junction trees purely in terms of observable variables, and to directly learn this alternative parametrization from observed quantities in training data. During learning, our algorithm first converts a latent graphical model into its corresponding latent junction tree, and then reduces the parameter learning problem down to a sequence of regression problems using a simple and fast approach called \textit{Two-Stage Regression}, which also allows us to incorporate prior knowledge into the learning process. Moreover, our proposed learning algorithm is also flexible enough to allow us to easily extend it to handle graphical models with continuous random variables using the technique of \textit{Hilbert Space Embeddings}. When dealing with continuous variables, we first embed their distributions into reproducing-kernel Hilbert spaces, and then use the kernel trick to perform all necessary learning operations over latent junction trees via tensor algebra.

The main contributions of our work are:

(1) We introduce a novel formulation of message-passing inference over latent-variable junction trees named \textbf{predictive belief propagation} (Section 3), and then propose a new algorithm for learning general latent-variable graphical models based on it using \textit{Two-Stage Regression} (Section 3 and 4). Our new algorithm overcomes many severe limitations faced by previous methods for learning latent graphical models, including EM and the spectral algorithms, and provides a general algorithmic framework that unifies the learning of all different kinds of latent graphical models.  

(2) We prove the correctness of our new algorithm by showing that it learns to compute a statistically consistent estimator of any conditional probability distribution over observable variables that we may query in a latent-variable graphical model during inference (Appendix \textbf{A.5}).

(3) We extend our algorithm from discrete domain to continuous domain using \textit{Hilbert Space Embeddings} of distributions (Section 5).

(4) We demonstrate that our learning algorithm outperforms both the EM algorithm and the spectral algorithm and runs significantly faster in experiments on both synthetic and real datasets (Section 6).

\subsection{Related Work}
Previously, Parikh et al. proposed a spectral algorithm \cite{Parikh12} for learning latent junction trees based on an alternative tensor parameterization. However, the algorithm in \cite{Parikh12} can only yield the marginal joint probability of all the observable variables in a latent graphical model together, without the ability to flexibly compute the posterior probability of arbitrary observable variables given other observed variables as evidence in a tractable manner. In contrast, our new algorithm overcomes this limitation and supports arbitrary inference in tractable forms by introducing a more flexible predictive message-passing paradigm. Moreover, our algorithm provides the ability to freely incorporate prior knowledge and to handle continuous variables, which also cannot be achieved by the spectral algorithm in \cite{Parikh12}.

\section{Formulation of the Learning Problem}
The central machine learning problem that we are dealing with in this paper is to learn general latent-variable probabilistic graphical models such that we can perform accurate inference over them among the observable variables. Traditionally, latent-variable graphical models are often parametrized using a set of local conditional probability tables (CPTs) that are associated with the edges in the graphs, and learning these models would mean to explicitly recover their CPT parameters from training data \cite{Koller09}. However, in most cases of application, the primary goal of learning a latent-variable graphical model is to be able to make accurate inference and predictions over its observable variables, and recovering its original CPT parameters is not needed at all. Therefore, in this paper we develop a new learning method that learns an alternative parametrization of general latent-variable graphical models purely based on observable quantities, such that we can directly perform accurate probabilistic inference on them using the learned alternative parametrization. We don't aim to learn the original CPT parameters of latent graphical models, and bypass them using our alternative parametrization.

For a latent-variable probabilistic graphical model $G$ of arbitrary graph structure, let $\mathscr{O}$ denote the set of all observable variables in $G$: $\mathscr{O} = \{ X_1, ..., X_{|\mathscr{O}|} \}$, and let $\mathscr{H}$ denote the set of all latent variables in $G$: $\mathscr{H} = \{ X_{|\mathscr{O}|+1}, ..., X_{|\mathscr{O}| + |\mathscr{H}|}\}$. Now our learning and inference problem at hand can be mathematically formulated as the following desired input-output behavior: 

\textbf{Input:} given a training dataset of $N$ $i.i.d.$ samples of the set of all observable variables $\{ x_1^d, ..., x_{|\mathscr{O}|}^d \}_{d=1}^{N}$, a set of observed evidence $\{ X_i = x_i \}_{i \in \mathcal{E}}$ (here $\mathcal{E}$ denotes the set of index for the set of observable variables that are observed as evidence), and the index $Q$ of the query node $X_Q$. \enspace 

\textbf{Output:} calculate an estimate of the posterior distribution of the query node conditioned on the observed evidence: $\widehat{\mathbb{P}}[X_Q \mid \{ X_i = x_i\}_{i \in \mathcal{E}}]$.

Here the variables in $G$ can be either discrete-valued or continuous-valued, and if the variables are continuous-valued, they are not restricted to have any specific functional forms of probability density function. Our algorithm can handle all of them gracefully in a nonparametric fashion.

\begin{figure}
  \centering
  \begin{tabular}{cc}
    \includegraphics[align = c, width=0.3\columnwidth]{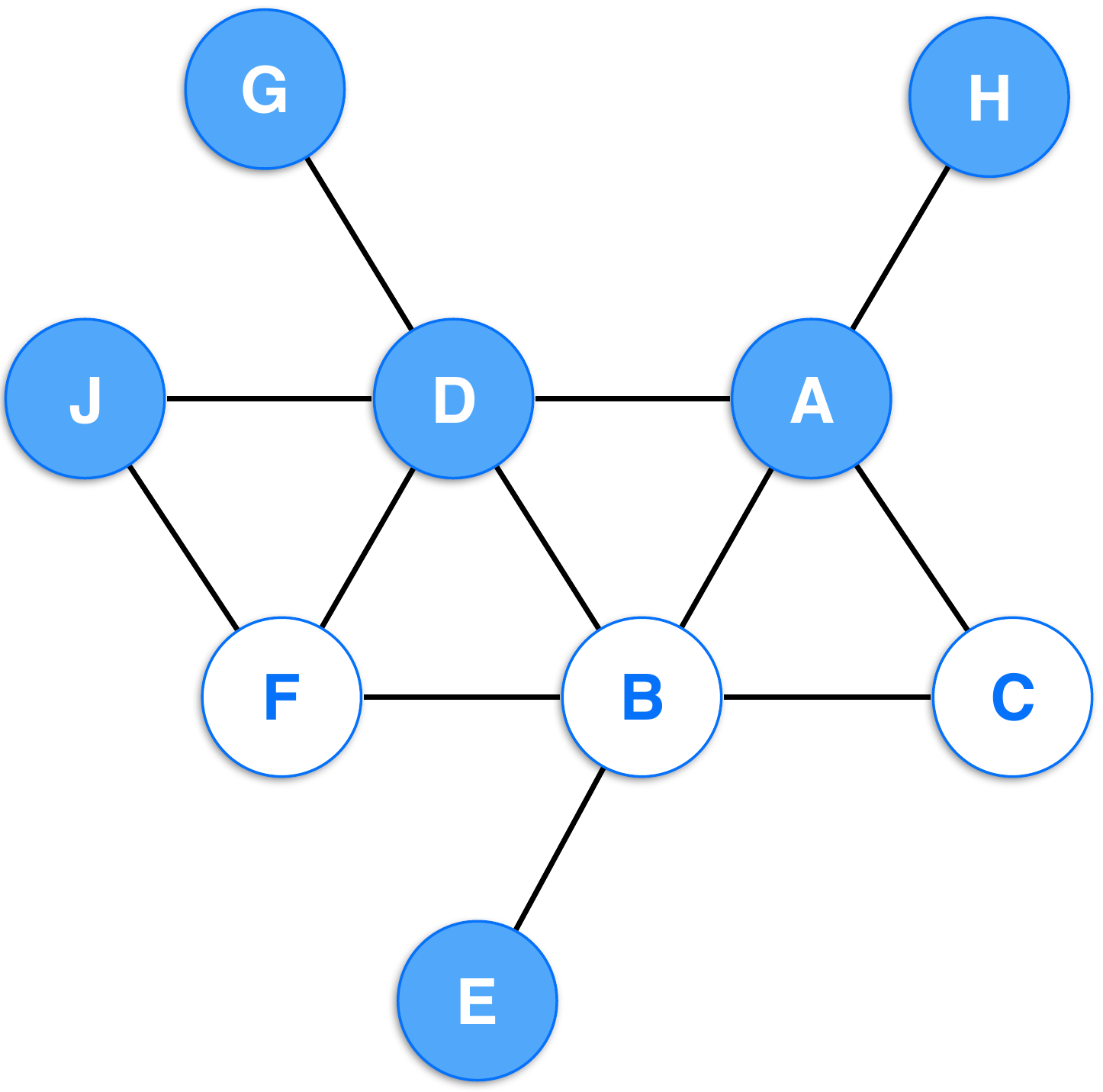}&
    \includegraphics[align = c, width=0.6\columnwidth]{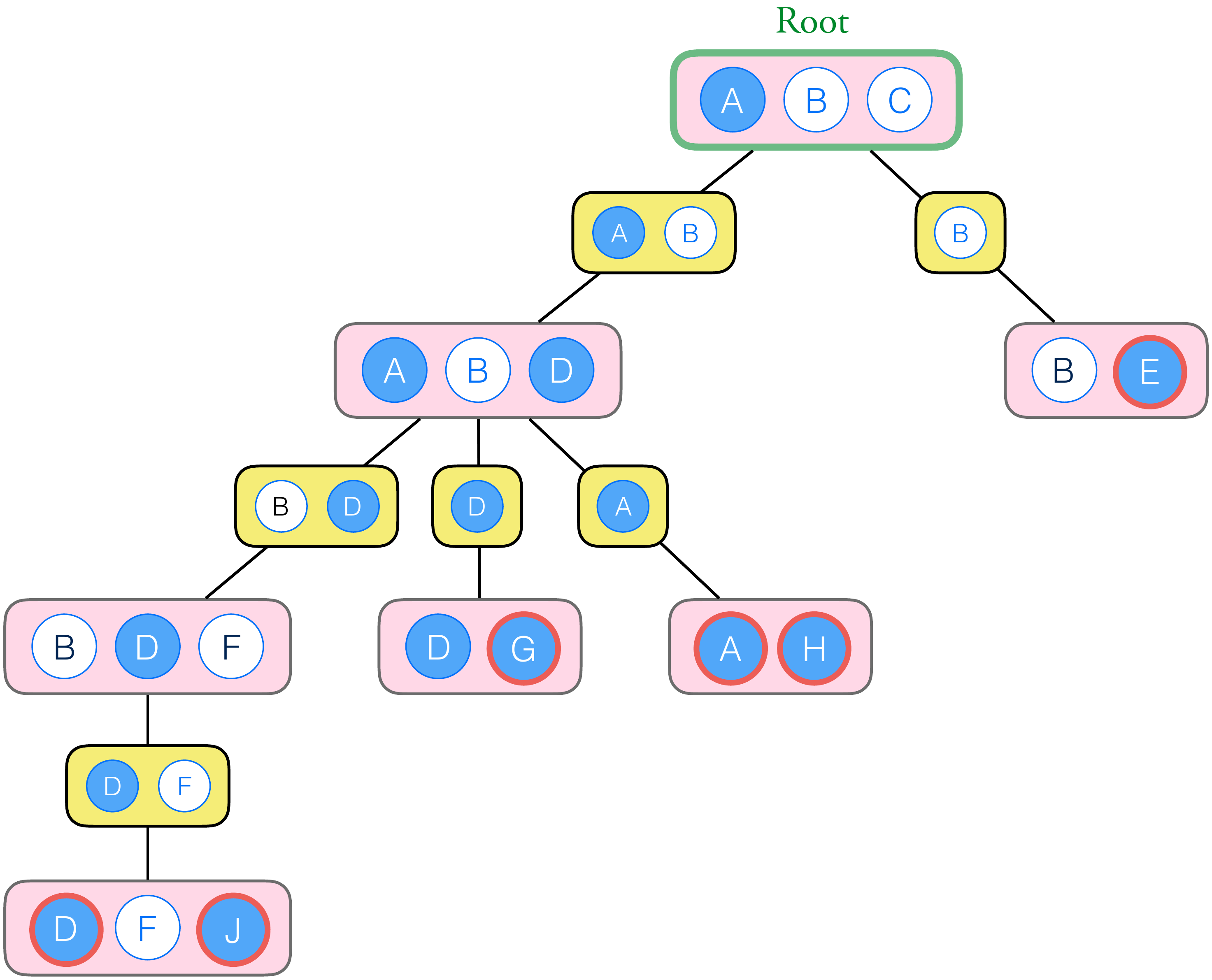}\\
    \makecell{(a)} & \makecell{(b)}
  \end{tabular}
  \caption{(a) An example of a latent-variable graphical model. Blue nodes indicate observable variables and white nodes indicate latent variables. (b) A corresponding latent junction tree that the graphical model in (a) is converted to. Pink squares indicate clique nodes and yellow squares indicate separator sets. Variable nodes with red circles around them are associated with their current leaf clique nodes.}
\end{figure}

\section{Predictive Belief Propagation}

\subsection{Preliminaries}
Here we first introduce two important notions that will be used throughout this paper.\\ 

\textbf{Sufficient Statistics Feature Vector} \quad In statistics and machine learning, the \textit{sufficient statistics feature vector} of a random variable $X$ is such a feature vector $v(X)$ that knowing its expectation $\mathbb{E}_{X \sim p}[v(X)]$ under $X$'s probabilistic distribution $p$ would completely determine that distribution $p$. And similarly, the \textit{sufficient statistics feature vector} of a group of random variables $\{X_1, ... , X_k\}$ is such a feature vector $v(\{X_1, ... , X_k\})$ that knowing its expectation $\mathbb{E}_{\{X_1, ... , X_k\} \sim p}[v(\{X_1, ... , X_k\})]$ under $\{X_1, ... , X_k\}$'s joint probabilistic distribution $p$ would completely determine that joint distribution $p$. For example, if $X$ is a discrete random variable, then its sufficient statistics feature vector $v(X)$ can be the vector of indicator functions whose $i$-th entry is 1 if $X$ takes the $i$-th value, and is 0 if otherwise. And for a group of discrete-valued random variables, their sufficient statistics feature vector can be the vectorized version of the outer product of all their individual vectors of indicator functions. See Appendix \textbf{A.4} for the continuous-valued case.\\

\textbf{Junction Trees} \quad In probabilistic graphical models, junction trees are a classical type of transformation that allows efficient message-passing inference algorithms to be performed over loopy graphical models. In order to design a learning and inference algorithm for general latent-variable graphical models, which may have loopy or non-loopy graph structures, we need to first resort to the junction tree algorithm \cite{Lauritzen88} to transform latent graphical models into their corresponding latent junction tree representations, over which we can perform message-passing inference. Consider the latent graphical model $G$ defined in Section 2. We first run the junction tree algorithm to convert $G$ into a latent junction tree $T$, and then associate each observable variable in $G$ with one leaf clique in $T$. Now pick a non-leaf clique node $C_r$ as the root of $T$, which naturally sets a topological order over $T$. Then for each separator set $S$ in $T$, we define its inside tree $In(S)$ to be the subtree rooted at $S$, and define its outside tree $Out(S)$ to be the rest of $T$ excluding $S$ and $In(S)$. See Figure 2 and 3 for an example.

\subsection{Difficulty Facing Conventional Belief Propagation}

In conventional methods for running belief propagation inference over junction trees, such as the Shafer-Shenoy algorithm \cite{Shenoy90} and the Hugin algorithm \cite{Lauritzen88,Anderson89}, the messages are defined to integrate together all the local probabilistic information from the past part of the junction tree in the form of partial results of sum-product calculations and send this compact summary of the past to the future part of the junction tree.\footnote{Formally, when we pass a message across a separator set in a junction tree, we can split the junction tree from this separator set into two separate subtrees. We refer to the subtree that the message-passing direction is pointing toward as the \textit{future} part of the junction tree, and the other subtree as the \textit{past} part of the junction tree. See Figure 3 for an example.} This requires the learning algorithm to be able to estimate the innate parametrization of the original latent graphical model, in the form of local conditional probability tables or local potential functions, directly from the training data. However, the innate parametrization of latent graphical models heavily involves hidden variables that we cannot observe, which makes it hard to directly learn this parametrization from training data. This discrepancy gives rise to the key difficulty in learning general latent graphical models, and forces previous methods to resort to inefficient local search heuristics such as EM.

\begin{figure}
  \centering
  \includegraphics[width=0.9\columnwidth]{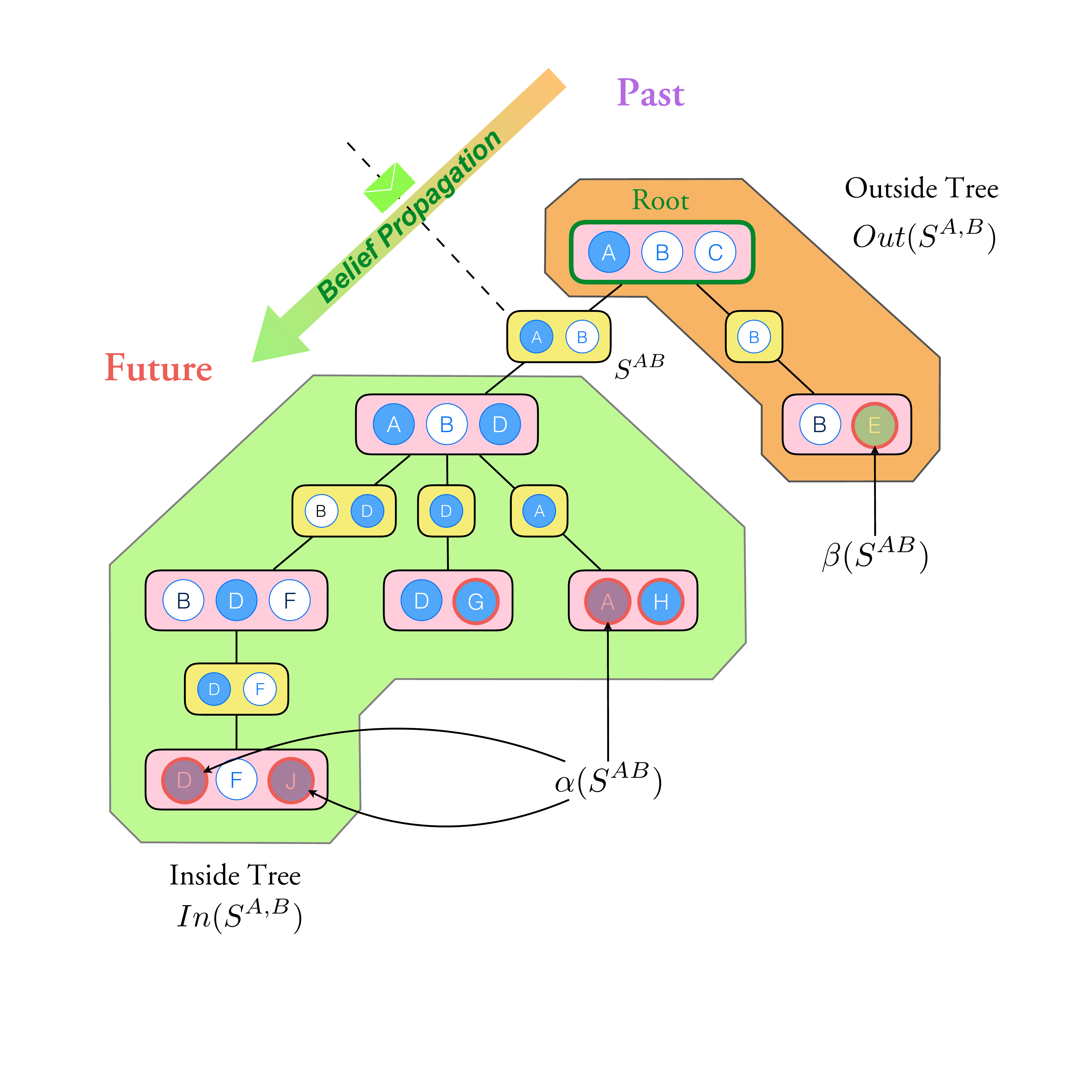}
  \caption{Illustration of the inside tree (future), outside tree (past), core group $\alpha(S^{AB})$ and evidence set $\beta(S^{AB})$ for a specific separator set $S^{AB}$ in the latent junction tree in Figure 2(b).}
\end{figure}

\subsection{The Central Idea of Predictive Belief Propagation}

To overcome this difficulty, we draw inspiration from the notion of \textit{predictive state representations} (PSRs) \cite{Littman01,Singh04}, which are a popular class of models for discrete-time dynamical systems. The key idea behind PSRs is to represent the state of a dynamical system as a set of predictions of features of observable variables in the future, which distinguishes them from the traditional history-based models and hidden-state-based models \cite{Kaelbling98}. Such an alternative representation of state based on observable quantities allows us to directly learn to perform filtering and prediction over dynamical systems from training data in a fast, provably consistent and local-optima-free fashion \cite{Boots11}.

In essence, the state of a system is a time bottleneck that compactly summarizes everything we need to know about the past in order to predict the future. From this viewpoint, when we perform belief propagation over a latent junction tree, the message that we send across a separator set can be viewed as the current state of the junction tree inference system at that particular separator set. Therefore, in analogy to PSRs, we can let the messages to encode our posterior predictions about the probabilistic information of the observable variables in the future part of the junction tree, given the observed evidence that we have absorbed from the past part of the junction tree. In PSRs, a common approach is to use a vector of sufficient statistics for a finite window of future observations to represent the state of a dynamical system. And analogously, in our case, at each separator set we can use a vector containing features of sufficient statistics for a subset of core observable variables in the future part of the junction tree to represent the message that we send across that separator set. Then during inference, we pass these predictive messages around over the latent junction tree to collect and propagate information we observe from evidence nodes and to compute the results of our inference queries. This forms our central idea of \textbf{predictive belief propagation}, which is the foundation of our new algorithm for learning latent-variable graphical models. The major advantage of this new way of thinking about message passing from a predictive perspective is that it enables us to efficiently learn general latent graphical models directly from observed quantities under a unified framework. Figure 1 provides an illustration of the comparison between predictive belief propagation and conventional belief propagation.

\subsection{Definition of Predictive Messages}

Now under the predictive belief propagation (PBP) framework, we first need to define what exactly are the predictive messages that we pass around over a latent junction tree during inference. As discussed above, the predictive messages in PBP are analogous to the predictive states in PSRs. In PSRs, a common approach is to use a vector of sufficient statistics (called the core set of tests) for a finite window of future observations to represent the predictive state of a dynamical system, because the infinite system-dynamics matrix can often be proven to have finite rank. Then analogously, in our PBP case, at each separator set we can use a sufficient statistics feature vector for a group of core observable variables in the future part of the junction tree to represent the predictive message that we send across that separator set. Now we make the following two key definitions:\\

\noindent \textbf{Definition 1.} We define the \textbf{core group of observable variables} $\alpha(S)$ for each separator set $S$ to be a subset of all the observable variables that are associated with the leaf clique nodes in $In(S)$ whose posterior joint distribution conditioned on evidence from $Out(S)$ completely determines the posterior joint distribution of all the observable variables that are associated with the leaf clique nodes in $In(S)$. More formally, let $\mathbf{OV}[In(S)]$ denote all the observable variables that are associated with the leaf clique nodes in the inside tree $In(S)$, and let $\mathbf{OV}[Out(S)]$ denote all the observable variables that are associated with the leaf clique nodes in the outside tree $Out(S)$, then the core group $\alpha(S)$ satisfies the conditional independence property that: $\{\mathbf{OV}[In(S)] \setminus \alpha(S)\} \CI \mathbf{OV}[Out(S)] \mid \alpha(S)$.

Such a core group of observable variables $\alpha(S)$ always exists for each separator set $S$, since at least the set $\mathbf{OV}[In(S)]$ would certainly qualify as being $\alpha(S)$, by definition. But in order to reduce the computational complexity of our learning and inference algorithm, it is desirable to find the minimal core group $\alpha(S)$ for each $S$. We present the process for determining the minimal core group in Appendix \textbf{A.2}.\\

\noindent \textbf{Definition 2.} Let $\theta^S[\alpha(S)]$ denote a sufficient statistic feature vector for $\alpha(S)$, and let $\Omega$ denote the evidence information that we observe from $Out(S)$. Then we define the \textbf{predictive message} that we send across each separator set $S$ to be the conditional expectation of $\theta^S[\alpha(S)]$ conditioned on $\Omega$, i.e. $\mathbb{E}\big[\theta^{S}[\alpha(S)] \mid \Omega \big]$. 

\subsection{Relationship between Predictive Messages}

Now the goal of our learning algorithm is to learn how these predictive messages relate to each other during PBP inference over a latent junction tree from training data. Without loss of generality, let's consider a non-leaf separator set $S$ in a latent junction tree $T$, where $S$ is connected with $K$ child separator sets $\{ S_1, S_2, ..., S_K\}$ below it through a clique node $C$. According to the definition of core groups in Definition 1, since $\alpha(S_1)$, $\alpha(S_2)$, ... , $\alpha(S_K)$ are all contained in $In(S)$, their posterior joint distributions $\mathbb{P}\big[\alpha(S_1) \mid \Omega \big], \mathbb{P}\big[\alpha(S_2) \mid \Omega \big], ..., \mathbb{P}\big[\alpha(S_K) \mid \Omega \big]$ are thus all completely determined by $\mathbb{P}\big[\alpha(S) \mid \Omega\big]$. Therefore, the conditional expectations $\mathbb{E}\big[\theta^{S_1}[\alpha(S_1)] \mid \Omega \big], \mathbb{E}\big[\theta^{S_2}[\alpha(S_2)] \mid \Omega \big], ..., \mathbb{E}\big[\theta^{S_K}[\alpha(S_K)] \mid \Omega \big]$, as well as their outer product, must also be fully determined by the conditional expectation $\mathbb{E}\big[\theta^S[\alpha(S)] \mid \Omega\big]$ (because all the $\theta$'s here are sufficient statistics feature vectors). That is to say, there exists a linear operator $\mathcal{W}^S$ in the form of a ($K$+1)-th order tensor with each mode corresponding to $S, S_1, S_2, ... , S_K$ respectively, such that:
\[\operatorname*{\otimes}\limits_{k=1}^K \mathbb{E}\big[\theta^{S_k}[\alpha(S_k)] \mid \Omega\big] = \mathcal{W}^S \times_{S} \mathbb{E}\big[\theta^S[\alpha(S)] \mid \Omega\big] \addtag \]
for any outside tree evidence $\Omega$, where $\otimes$ denotes outer product and $\times_S$ denotes mode-specific tensor multiplication\footnote{For a detailed introduction to tensor algebra, we refer the readers to \cite{Kolda09}. In this paper we adopt the notations from \cite{Parikh12} to label the modes of tensors with random variables.} along mode $S$.

Therefore, the predictive messages in PBP relate to each other through $\mathcal{W}^S$ according to Eq. (1). This linear operator $\mathcal{W}^S$ essentially acts as a message processor and distributor during PBP inference.

\subsection{Two-Stage Regression on Latent Junction Trees}

Therefore, the major goal of our learning algorithm is to learn the linear operator $\mathcal{W}^S$ for each non-leaf separator set $S$ in $T$ from training data. Previously, \cite{Hefny15} proposed a method named \textit{Two-Stage Regression} (2SR) to learn PSRs of linear dynamical systems. 2SR learns PSRs by solving a sequence of regression problems, and is fast and statistically consistent. The key idea of 2SR is to use instrumental variable regression \cite{Stock11} to recover an unbiased estimate of the linear mappings in PSRs. Here we generalize 2SR to latent junction trees to learn $\mathcal{W}^S$. First, pick a feature vector for $Out(S)$ and denote it by $\eta^S[Out(S)]$. Then using $\eta^S[Out(S)]$ as our instrumental variable, we can perform our \textit{Two-Stage Regression} to learn $\mathcal{W}^S$ in three steps: (1) regress $\theta^S[\alpha(S)]$ on $\eta^S[Out(S)]$; (2) regress $\operatorname*{\otimes}\limits_{k=1}^K \theta^{S_k}[\alpha(S_k)]$ on $\eta^S[Out(S)]$; (3) run a linear regression from the predictions obtained in step (1) to the predictions obtained in step (2) to recover an unbiased estimate of $\mathcal{W}^S$. These three steps of supervised learning constitute our new \textit{Two-Stage Regression} approach for latent junction trees, which will manifest itself in the learning algorithm in Section 4.

\section{Main Algorithm}

We are now ready to present our new algorithm for learning general latent-variable probabilistic graphical models based on the framework of predictive belief propagation. It has two components: a learning algorithm and a corresponding inference algorithm that allows us to perform probabilistic inference over latent graphical models using the results learned from the learning algorithm. Here we first describe the basic version of our algorithm for the case where all observable variables in a graphical model are discrete-valued. Then we show how to extend our algorithm to handle graphical models with continuous-valued variables in Section 5. See Appendix \textbf{A.3} for a pseudocode summary of our algorithm and see Appendix \textbf{A.5} for the proof of consistency of our algorithm.

\subsection{The Learning Algorithm}

\quad \textbf{Step 1. Model Construction:}
Run the junction tree algorithm to convert $G$ into an appropriate latent-variable junction tree $T$ and pick a root $C_r$ for $T$, such that each observable variable in $G$ can be associated with one leaf clique in $T$. See Figure 2(a) and 2(b) for a concrete example.\\

\textbf{Step 2. Model Specification:} 
For each separator set $S$ in $T$, among all the observable variables that are associated with the leaf clique nodes in its inside tree $In(S)$, determine its minimal core group $\alpha(S) = \{A_1, A_2, ... , A_{|\alpha(S)|}\}$ using the procedure described in Appendix \textbf{A.2}. And among all the observable variables associated with its outside tree $Out(S)$, select a subset of variables $\beta(S) = \{B_1, B_2, ... , B_{|\beta(S)|}\}$ (this can be any subset). (See Figure 3 for a concrete example.) Now pick a feature vector $\theta^S[\alpha(S)]$ for $\alpha(S)$ and a feature vector $\eta^S[\beta(S)]$ for $\beta(S)$, where we require that $\theta^S$ must be a sufficient statistics feature vector for $\alpha(S)$ and $\eta^S$ can be any feature vector. For discrete-valued $\alpha(S)$, this sufficient statistic feature vector can simply be the (vectorized) outer product of the vectors of indicator functions of all the variables in $\alpha(S)$; and for continuous-valued $\alpha(S)$, this sufficient statistic feature vector can be the (vectorized) outer product of the implicit feature map of characteristic kernels of all the variables in $\alpha(S)$ (see Appendix \textbf{A.4}).\\

\textbf{Step 3. Stage 1A Regression (S1A):} 
At each non-leaf separator set $S$ in $T$, learn a (possibly non-linear) regression model to estimate $\bar{\theta^S} = \mathbb{E}[\theta^S \mid \eta^S]$. The training data for this regression model is $\big\{\big(\theta^S[\alpha(S)^d], \eta^S[\beta(S)^d]\big)\big\}_{d=1}^N$ across all $N$ $i.i.d.$ training samples.\\

\textbf{Step 4. Stage 1B Regression (S1B):} 
At each non-leaf separator set $S$ in $T$, where $S$ is connected with $K$ child separator sets $\{ S_1, S_2, ..., S_K\}$ below it, learn a (possibly non-linear) regression model to estimate $\bar{\operatorname*{\otimes}\limits_{k=1}^K \theta^{S_k}} = \mathbb{E}[\operatorname*{\otimes}\limits_{k=1}^K \theta^{S_k} \mid \eta^S]$. The training data for this regression model are $\big\{\big(\operatorname*{\otimes}\limits_{k=1}^K \theta^{S_k}[\alpha(S_k)^d], \eta^S[\beta(S)^d]\big)\big\}_{d=1}^N$ across all $N$ $i.i.d.$ training samples.

\textit{Note}: In the S1A and S1B regression steps above, we can use any supervised learning algorithm as our regression model. This provides us with the flexibility to incorporate different prior knowledge into our learning process.\\

\textbf{Step 5. Stage 2 Regression (S2):}
At each non-leaf separator set $S$ in $T$, use the feature expectations estimated in S1A and S1B to train a linear regression model to predict $\bar{\operatorname*{\otimes}\limits_{k=1}^K \theta^{S_k}} = \mathcal{W}^S \times_{S} \bar{\theta^S}$, where $\mathcal{W}^S$ is the linear operator associated with $S$. Output the learned parameter tensor $\mathcal{W}^S$. The training data for this linear regression model are estimates of $\big( \bar{\operatorname*{\otimes}\limits_{k=1}^K \theta^{S_k}}, \bar{\theta^S} \big)$ for all the training samples that we obtained from S1A and S1B regressions.\\

\textbf{Step 6. Root Tensor Estimation:}
At the root $C_r$, estimate the expectation of the outer product of the inside tree feature vectors of all adjacent separator sets that are connected with $C_r$ by taking average across all the $N$ $i.i.d.$ training samples: $\mathcal{T}^{C_r} = \widehat{\mathbb{E}}\big[\operatorname*{\otimes}\limits_{S \in \gamma(C_r)} \theta^S[\alpha(S)]\big] = \dfrac{1}{N} \sum\limits_{d = 1}^{N} \operatorname*{\otimes}\limits_{S \in \gamma(C_r)} \theta^S[\alpha(S)^d]$ where $\gamma(C_r)$ denotes the set of all separator sets that are connected to $C_r$. Output the learned parameter tensor $\mathcal{T}^{C_r}$. This root tensor $\mathcal{T}^{C_r}$ will later serve the function of exchanging information at the root clique $C_r$ during Step 3(2) of the inference algorithm below.\\

\textbf{Output:} The final outputs of our learning algorithm are the linear operators $\mathcal{W}^S$ for each non-leaf separator set that we obtained from Step 5 above, and the root tensor $\mathcal{T}^{C_r}$ that we obtained from Step 6 above. These $\mathcal{W}^S$ and $\mathcal{T}^{C_r}$ essentially serve as an alternative parametrization of the latent-variable graphical model $G$ that supports inference based on \textit{predictive belief propagation}. In our inference algorithm below, we will use $\mathcal{W}^S$ and $\mathcal{T}^{C_r}$ to build our message-passing protocols and to calculate the result of our inference query.

\subsection{The Inference Algorithm}

Our inference algorithm uses the learned alternative parametrization $\mathcal{W}^S$ and $\mathcal{T}^{C_r}$ outputed by the learning algorithm above to compute inference query through \textit{predictive belief propagation}.\\

\textbf{Step 1. Leaf Tensor Construction:}
For each leaf separator set $S_l$ of $T$, whose core group is $\alpha(S_l) = \{A_1, A_2, ... , A_{|\alpha(S_l)|}\}$, construct a leaf tensor $\Phi^{S_l}$ with modes $\theta^{S_l}$, $A_1$, $A_2$, ..., $A_{|\alpha(S_l)|}$ and dimensions $length(\theta^{S_l}) \times \mathcal{N}(A_1) \times \mathcal{N}(A_2) \times ... \times \mathcal{N}(A_{|\alpha(S_l)|})$, where $\mathcal{N}(X)$ denotes the number of values for a discrete random variable $X$. Each fiber\footnote{As explained in \cite{Kolda09}, a fiber of a tensor is a higher-order analogue of matrix rows and columns. A fiber along a mode is defined by fixing every mode of the tensor except for that one.} of the tensor $\Phi^{S_l}$ along mode $\theta^{S_l}$ takes the value of the feature vector $\theta^{S_l}$ evaluated at the corresponding values of variables $A_1, A_2, ... , A_{|\alpha(S_l)|}$, i.e., 
\begin{center}
$\quad \Phi^{S_l}\big(:,\mathcal{I}(a_1),\mathcal{I}(a_2),...,\mathcal{I}(a_{|\alpha(S_l)|})\big) = \theta^{S_l}[a_1,a_2,...,a_{|\alpha(S_l)|}]$
\end{center}
where $\mathcal{I}(a)$ denotes the order index of a value $a$ in all the possible values of $A$. The purpose of these leaf tensors $\Phi^{S_l}$ is to completely encode the sufficient statistic feature information of all the observable variables in $G$ into our PBP inference system.\\

\textbf{Step 2. Initial Leaf Message Generation:}
At each leaf clique node $C_l$ in $T$, let $\delta(C)$ denote the set of all observable variables that are associated with $C_l$, and let $S_l$ denote the separator set right above $C_l$. We define a function $\zeta(X)$ of an observable variable $X$ that evaluates to an all-one vector if $X$ is not observed in the evidence and evaluates to a one-hot-encoding value indicator vector $e_x$ if $X$ is observed to have value $x$ in the set of observed evidence\footnote{For example, if $X$ is a discrete random variable that can take 6 possible values $\{1,2,3,4,5,6\}$, and X is observed to have value 3 in the evidence, then $e_{3} = [0,0,1,0,0,0]$}. That is to say:
\[\zeta(X) = \begin{cases}
      \vec{1}, & \text{if } X \text{ is not observed} \\
      e_x, & \text{if } X \text{ is observed to have value } x
    \end{cases}\]
Then the upward message that we send from $C_l$ to its parent clique node $C_p$ can be calculated as:
\[ m_{C_l \rightarrow C_p} = \Phi^{S_l^{\dagger}} \times_{\{\delta(C)\}} \big[\operatorname*{\otimes}\limits_{X \in \delta(C)} \zeta(X)\big] \addtag \]
where ${\dagger}$ denotes Moore-Penrose pseudoinverse. This step serves the purpose of collecting all the observed information in the evidence.\\

\textbf{Step 3. Message Passing:}

(1) \textit{From leaf to root (the upward phase):} \enspace For each non-root parent clique node $C$ in $T$ where it is separated by a separator set $S$ from its parent node $C_p$, once it has received the messages from all of its child nodes $C_1, C_2, ..., C_K$, which are separated from it by separator sets $S_1, S_2, ..., S_K$ respectively, compute and send the upward message:
\[ m_{C \rightarrow C_p} = \mathcal{W}^S \operatorname*{\times_{S_{j}}}\limits_{j \in \{1,2,...,K\}} m_{C_j \rightarrow C} \addtag \]
This step gradually collects all the local observed evidence information from all the leaf cliqes up to the root clique. 

(2) \textit{At the Root:} \enspace At the root clique node $C_r$ of $T$, which is surrounded by its $K$ child nodes $C_1,...,C_K$ (each separated from it by separator sets $S_1,...,S_K$ respectively), for each $k \in \{1,2,...,K\}$, once $C_r$ has received all the $K-1$ upward messages from all of its other $K-1$ child nodes except for $C_k$, compute and send the downward message:
\[ m_{C_r \rightarrow C_k} = \mathcal{T}^{C_r} \operatorname*{\times_{S_{j}}}\limits_{j \in (\{1,2,...,K\} \backslash k)} m_{C_j \rightarrow C_r} \addtag \]
This step summarizes and exchanges evidence information from different subtrees at the root clique.

(3) \textit{From root to leaf (the downward phase):} \enspace For each non-root parent clique node $C$ in $T$ where it is separated from its parent node $C_p$ by a separator set $S$ and separated from its $K$ child nodes $C_1, C_2, ..., C_K$ by separator sets $S_1, S_2, ..., S_K$ respectively, once it has received the downward message $m_{C_p \rightarrow C}$ from $C_p$, for each $k \in \{1,2,...,K \}$, compute and send the downward message:
\[m_{C \rightarrow C_k} \nonumber = \mathcal{W}^S \times_{S} m_{C_p \rightarrow C} \operatorname*{\times_{S_{j}}}\limits_{j \in (\{1,2,...,K\} \backslash k)} m_{C_j \rightarrow C} \addtag \]
This step is the core operation of \textit{predictive belief propagation}. We gradually compute predictive messages level by level from the root clique down to the leaf cliques. All the downward messages $m_{C \rightarrow C_k}$ in this step are \textit{predictive messages}, as defined in Definition 2.\\

\textbf{Step 4. Computing Query Result:}
For the query node $X_Q$ associated with $C_Q$, denote $C_Q$'s parent node as $C_p$ and the separator set between them as $S_Q$. First use the leaf tensor $\Phi^{S_Q}$ and its Moore-Penrose pseudoinverse $\Phi^{S_Q^{\dagger}}$ to transform the downward incoming message $m_{C_p \rightarrow C_Q}$ and the upward outgoing message $m_{C_Q \rightarrow C_p}$, respectively, and then compute the Hadamard product of these transformed versions of the two messages to obtain an estimate of the unnormalized conditional probability of $\delta(C_Q)$ given all the evidence $\{X_i = x_i\}_{i \in \mathcal{E}}$:
\begin{center}
$\widehat{\mathbb{P}}[\delta(C_Q) \mid \{ X_i = x_i \}_{i \in \mathcal{E}}] \propto \big(m_{C_p \rightarrow C_Q} \times_{S_Q} \Phi^{S_Q^{\dagger}} \big) \circ \big( \Phi^{S_Q} \times_{S_Q} m_{C_Q \rightarrow C_p}  \big)$
\end{center}

Now we marginalize out the variables in $\delta(C_Q) \backslash X_Q$ and renormalize to obtain the final query result - the estimate of the conditional probability distribution of the query variable $X_Q$ given all the evidence: $\widehat{\mathbb{P}}[X_Q \mid \{ X_i = x_i\}_{i \in \mathcal{E}}]$.\\

\textbf{[Additional Note]}: Another important type of query that we can also compute here is the joint probability of all the observed evidence: $\widehat{\mathbb{P}}[ \{ X_i = x_i\}_{i \in \mathcal{E}}]$. In Step 4 above, before marginalization and renormalization, the Hadamard product is indeed equal to $\widehat{\mathbb{P}}[\delta(C_Q), \{ X_i = x_i \}_{i \in \mathcal{E}}]$ (see Appendix \textbf{A.5} for the proof). We can marginalize out all the variables in $\delta(C_Q)$ from it to obtain $\widehat{\mathbb{P}}[ \{ X_i = x_i\}_{i \in \mathcal{E}}]$. For example, this type of query is used for classification in the handwritten digit recognition task in the experiment in Section 6.

\section{Extending to Continuous Domain through RKHS Embeddings}
One of the biggest advantages of our new algorithm compared to previous methods is that it can be seamlessly extended from discrete domain to continuous domain in a nonparametric fashion using the technique of kernel embeddings. When encountering continuous random variables in a latent graphical model, we use reproducing-kernel Hilbert space (RKHS) embeddings \cite{Boots13,Song10b,Song10,Song11b} of distributions as sufficient statistic features and express all the learning and belief propagation operations as tensor algebra in the infinite-dimensional Hilbert space, and then employ the kernel trick to transform these operations back into tractable finite-dimensional linear algebra calculations over Gram matrices \cite{Song11,Grune12}. We present our full derivation of this extension in Appendix \textbf{A.4}.

\section{Experiments}
We design two sets of experiments to evaluate the performance of our proposed algorithm, one using synthetic data and the other one using real data. 

\begin{figure}[h]
    \centering
    \includegraphics[align = c, width=0.4\columnwidth]{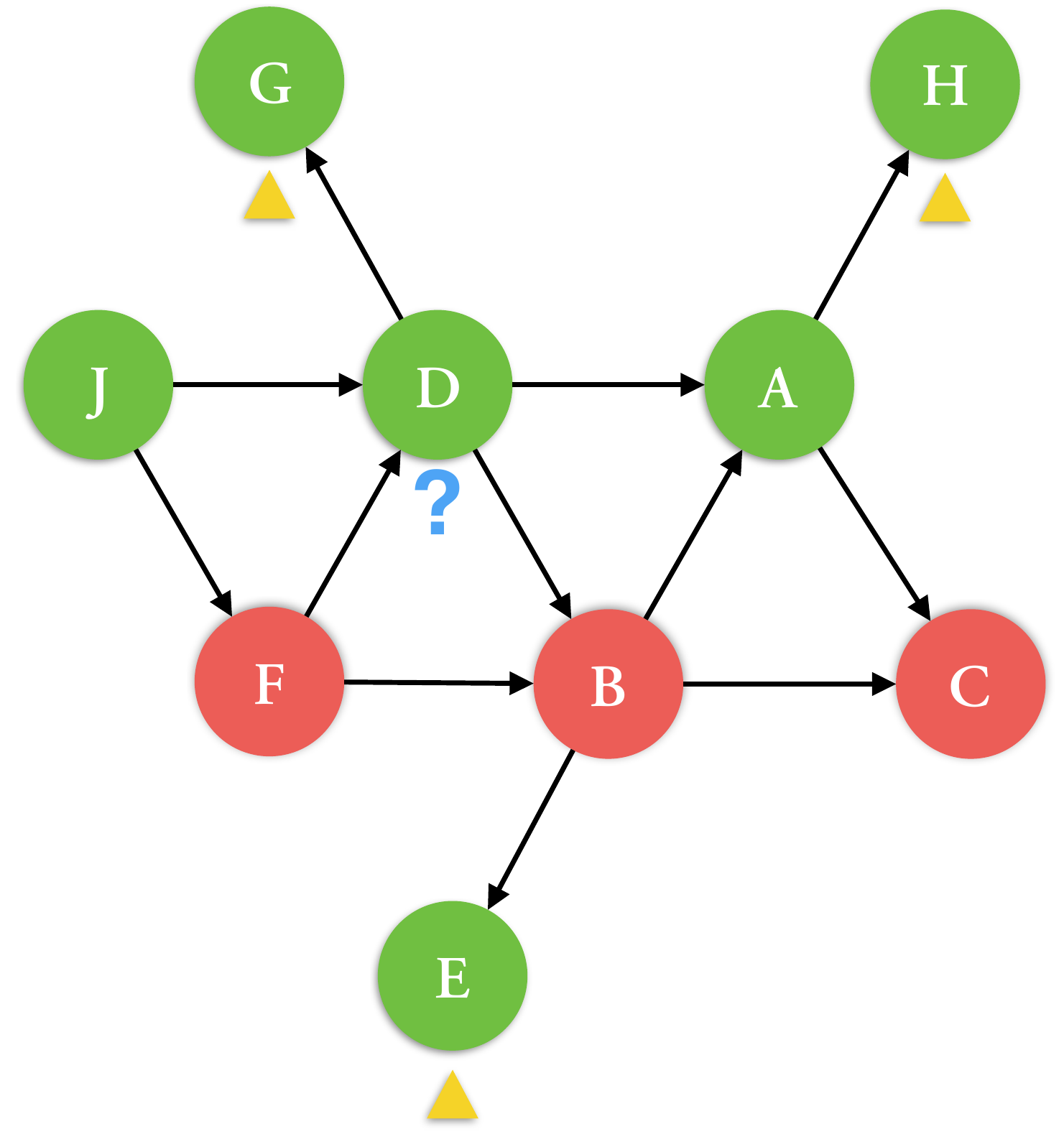}
    \caption{A directed latent-variable graphical model. Green nodes indicate observable variables and red nodes indicate latent variables.}
\end{figure}

\subsection{Synthetic Dataset}
In this experiment, we test the performance of our algorithm on the task of learning and running inference on the discrete-valued latent-variable graphical model depicted in Figure 4 using artificially generated synthetic data and compare it with both the standard EM algorithm \cite{Dempster77} and the stepwise online EM algorithm \cite{Liang09}. 

\begin{figure}[h]
\centering
\includegraphics[align = c, width = 0.7\columnwidth]{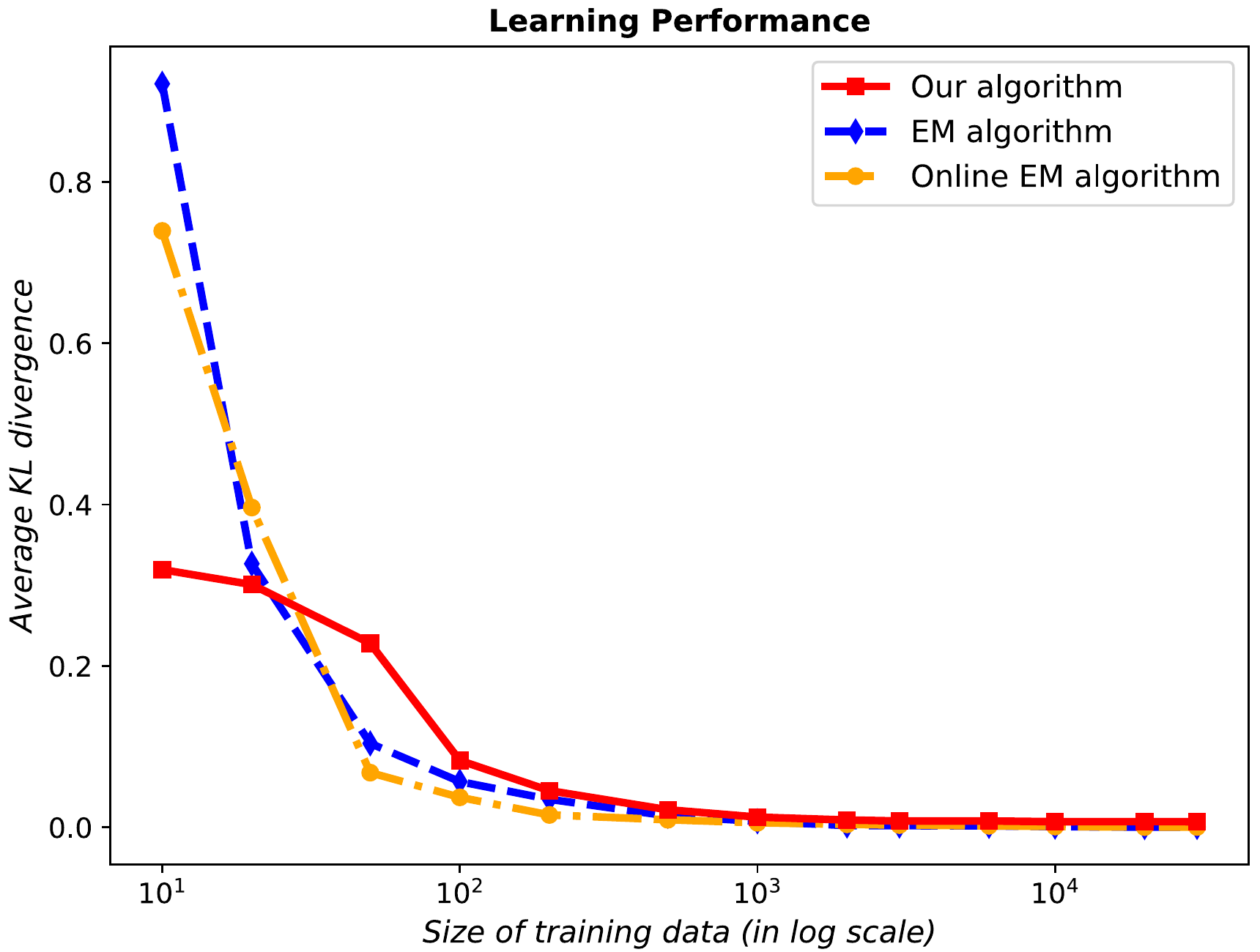}
\\
\quad\\
\quad\\
\includegraphics[align = c, width = 0.7\columnwidth]{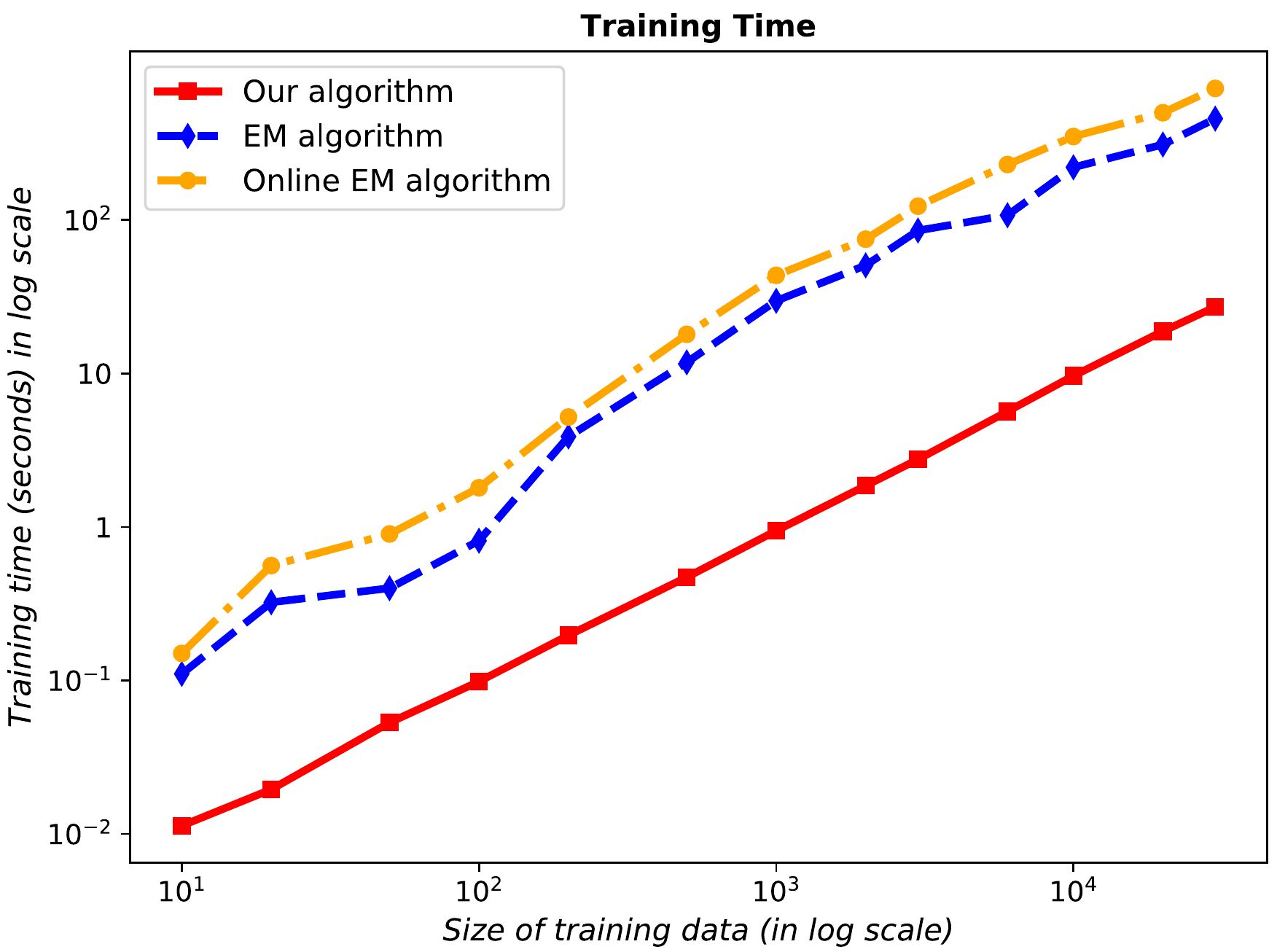}
\caption{Comparison between our learning algorithm, the EM algorithm and the stepwise online EM algorithm on the synthetic dataset.}
\end{figure}

We randomly initialize all the conditional probability tables in this model as the ground truth parameters, and then sample a dataset containing joint observations of the observable variables. Next we apply our proposed algorithm to learn this model, and evaluate its performance on the task of inferring the posterior distribution of variable $D$ given the observed values at variables $G$, $H$, and $E$. In our experiment, we use ridge regression \cite{Friedman06} for S1A and S1B. We compute the Kullback-Leibler divergence between our algorithm's inferred posterior and the ground truth posterior calculated using the exact Shafer-Shenoy algorithm and average across all possible joint realizations of the variables $(g,h,e)$. We report the results in Figure 5, where we see that the average KL divergence between our algorithm's results and the ground truth posterior quickly decreases and approaches 0 as the size of the training data increases. This result demonstrates that our algorithm learns quickly to perform accurate inference over latent graphical models.

We also run the standard EM algorithm \cite{Dempster77} and the stepwise online EM algorithm \cite{Liang09} to learn the same model with the same synthetic dataset, and compare their performance and training time with our algorithm (Figure 5).\footnote{In our experiment, we give both EM and online EM 10 random restarts and take the best ones as their performance evaluation.} Our algorithm achieves equally good learning performance as EM and online EM do, but is much faster to train than both of them. The spectral algorithms can not perform such inference task on individual observable variables in a tractable manner, so we didn't include them as our baselines here.

\begin{figure}[h]
\centering
\includegraphics[align = c, width=0.8\columnwidth]{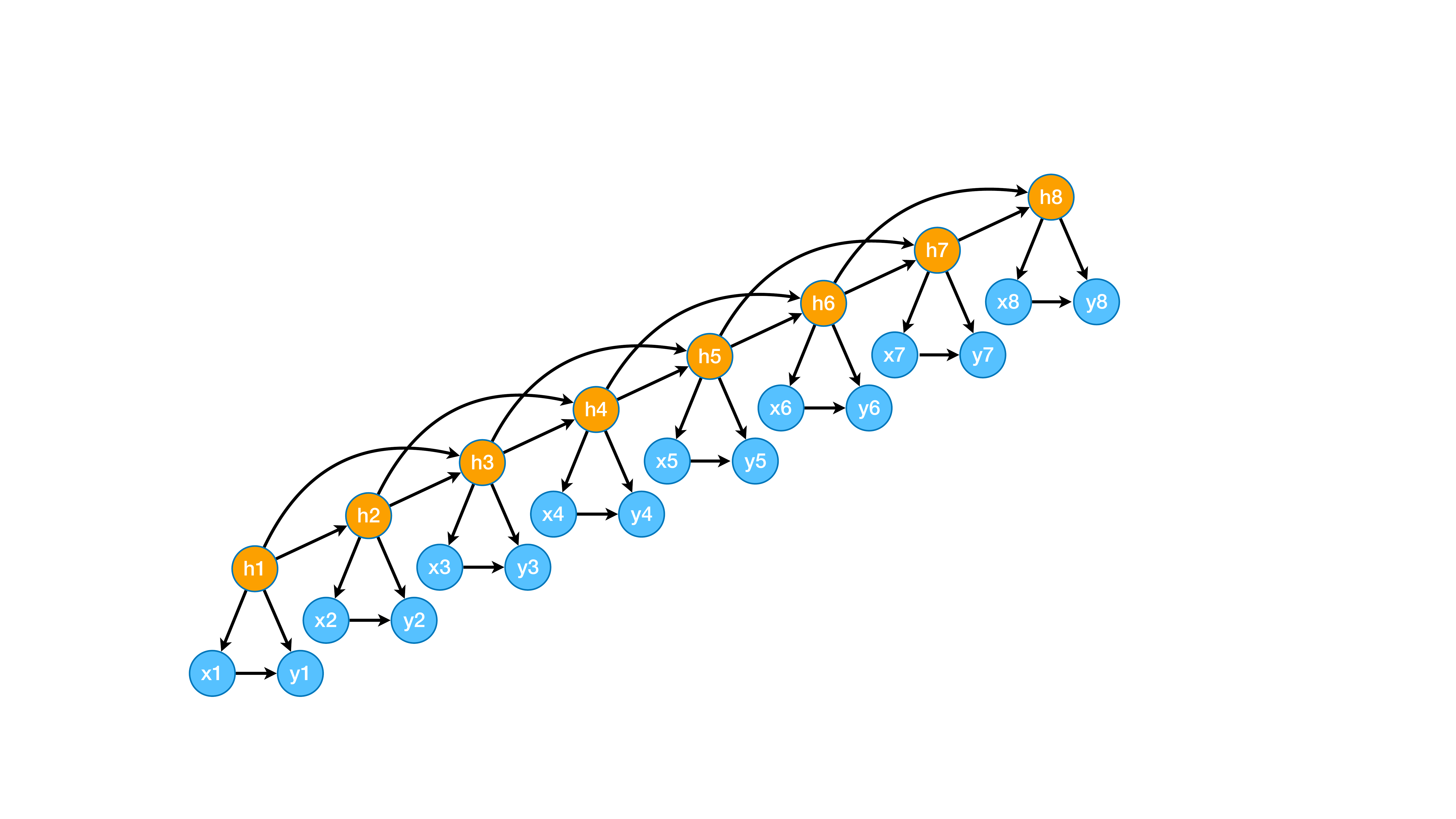}
\caption{A directed latent-variable graphical model to model the generative process of the 16-dimensional feature vectors of handwritten digits.}
\end{figure}

\subsection{Handwritten Digit Recognition}
In this experiment we consider the task of recognizing handwritten digits using the Pen-Based Recognition of Handwritten Digits dataset in the UCI machine learning repository \cite{UCI}. This dataset collected 10992 handwritten digit samples from 44 writers on a tablet with $500\times500$ pixel resolution and then normalized all the coordinates into integer values between 0 and 100. It then used spatial resampling to obtain 8 regularly spaced points to represent each handwritten digit, and the feature vector for each digit is the 16-dimensional vector consisting of the $(x,y)$ coordinates of the 8 representative points. In order to learn to classify these handwritten digits, we design a latent variable graphical model structure (shown in Figure 6) to model the generative process of the 16-dimensional feature vectors of handwritten digits. The blue nodes indicate observable variables that corresponds to the coordinate values, and the orange nodes indicate latent variables. We apply our learning algorithm to learn a different generative model for each of the 10 digit categories, and then during test time, we use our inference algorithm to calculate the probability that a test instance is generated from each of the 10 different models, and choose the one with the highest probability as our predicted category. Here we use 7000 samples as our training set, 494 samples as our validation set, and the other 3498 samples as our testing set. In our experiment, we use Gaussian radial basis function kernel embeddings with bandwidth parameter $\sigma = 10$ as our feature vectors, and use ridge regression \cite{Friedman06} with regularization parameter $\lambda = 0.1$ for S1A and S1B.

\begin{figure}[h]
\begin{center}
\includegraphics[align = c, width=0.8\columnwidth]{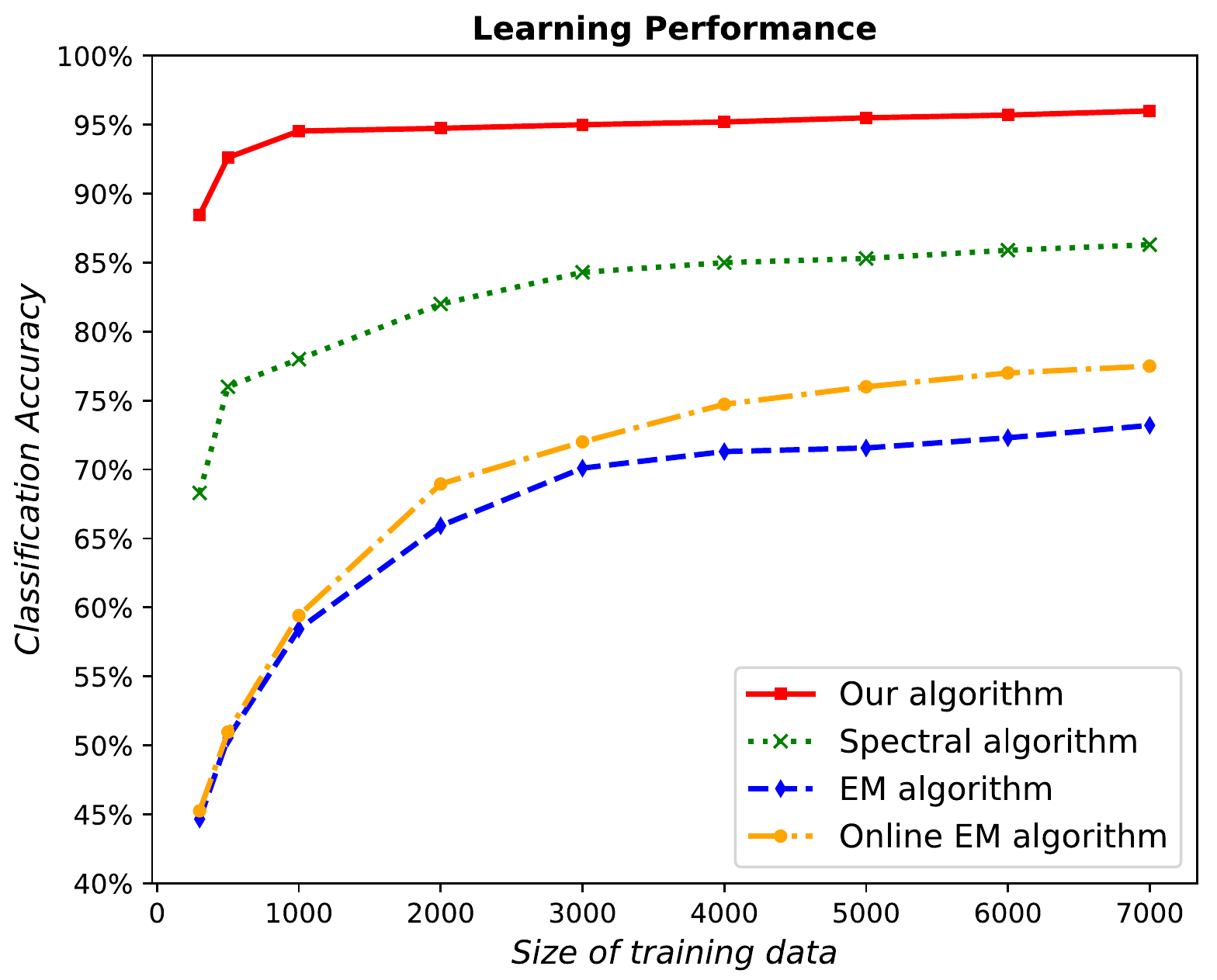}\\
\quad\\
\quad\\
\includegraphics[align = c, width=0.8\columnwidth]{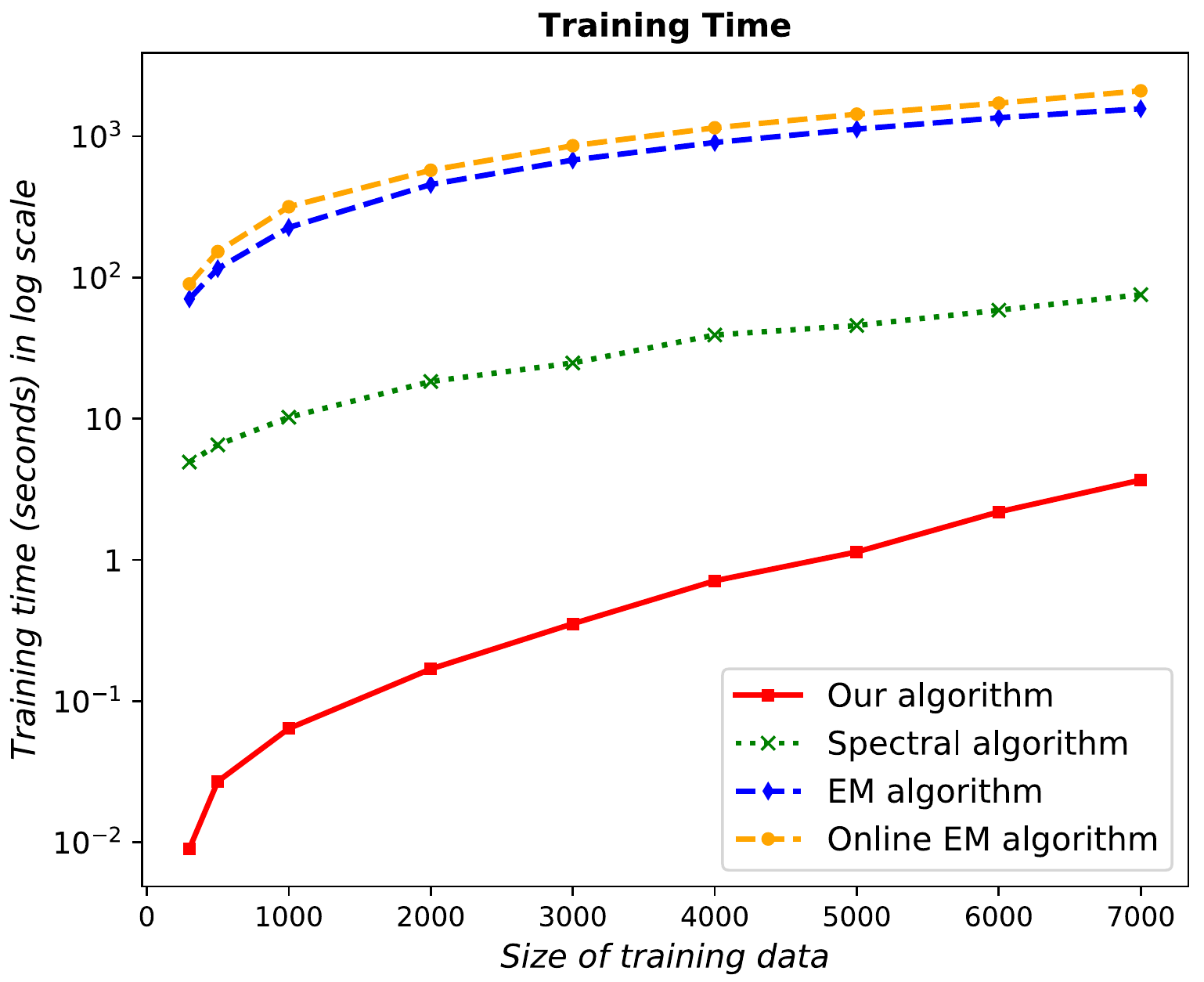}
\caption{Comparison of our learning algorithm, the spectral algorithm, the EM algorithm and the stepwise online EM algorithm on the handwritten digit recognition experiment.}
\end{center}
\end{figure}

We also run the standard EM algorithm \cite{Dempster77}, a stepwise online EM algorithm \cite{Liang09}, and a spectral algorithm as the baselines to learn the same model, and compare their classification accuracy and training time with our algorithm (Figure 7). From the experimental results we can clearly see that our learning algorithm performs much better and is much faster to train than both the spectral algorithm and the two EM algorithms on this handwritten digit classification task. And moreover, we observe that our algorithm is also very robust and yields good performance even when the size of the training data is relatively small, while the other three algorithms all perform poorly in this scenario. 

\section{Conclusion}
In this paper, we have introduced predictive belief propagation as a new formulation of message-passing inference over latent junction trees and developed a new algorithm for learning general latent-variable graphical models based on it. Our new algorithm unifies the learning and inference of all different types of latent graphical models under a single flexible framework, and overcomes many severe limitations faced by previous methods like EM and spectral algorithms (see Appendix \textbf{A.1} for a detailed comparison between our algorithm and previous algorithms). We also proved that our new algorithm gives a consistent estimator of inference queries over all latent graphical models. We evaluated its performance on both synthetic and real datasets, and showed that it learns different types of latent graphical models efficiently and achieves superior inference performance. Therefore, we believe that our algorithm provides a powerful and flexible new learning framework for general latent-variable graphical models.

\bibliographystyle{aaai}
\bibliography{paper}

\end{document}